\documentstyle{article}

\begin{document}

\begin{center}
{\Large {\bf Natural Language Generation in Healthcare}}

\medskip
{\large Brief Review - to appear in {\bf JAMIA}, 1997}

\bigskip
Alison J. Cawsey, M.Sc., Ph.D.$^1$   Bonnie L. Webber, Ph.D.$^2$    Ray B. Jones, M.Sc., Ph.D.$^3$
\end{center}

\vspace*{1in}
\begin{abstract}
Good communication is vital in healthcare, both among healthcare
professionals, and between healthcare professionals and their patients.
And well-written documents, describing and/or 
explaining the information in structured databases may be easier to comprehend, more edifying 
and even more convincing, than the structured data, even when presented in tabular or graphic 
form. Documents may be automatically generated from structured data, using techniques from 
the field of natural language generation. These techniques are concerned with how the content, 
organisation and language used in a document can be dynamically selected, depending on the 
audience and context. They have been used to generate health education materials, 
explanations and critiques in decision support systems, and medical reports
and progress notes.
\end{abstract}


\vspace*{0.5in}
\begin{flushleft}
Reprint requests and communication should be sent to: \\
Alison Cawsey\\
Department of Computing and Electrical Engineering\\
Heriot-Watt University\\
Edinburgh EH14 4AS\\
Scotland\\

email: {\tt alison@cee.hw.ac.uk}\\
tel: +44-131-451-3413\\
fax: +44-131-451-3327
\end{flushleft}

\vspace*{0.5in}
\noindent
$^1$ Department of Computing and Electrical Engineering, Heriot-Watt
University, Edinburgh EH14 4AS, Scotland

\smallskip
\noindent
$^2$ Department of Computer and Information Science, University of
Pennsylvania, Philadelphia, PA 19104-6389, USA

\smallskip
\noindent
$^3$ Department of Public Health, University of Glasgow, Glasgow G12 8RZ, Scotland.

\newpage
\section{Introduction}

Healthcare is an area where effective communication is vital, both between healthcare 
providers and their patients, and among healthcare providers themselves. Different participants 
in the healthcare process - consultants, nurses, general practitioners, medical researchers, 
patients, theirrelatives, and even accountants and administrators - must all be able to obtain 
and communicate relevant information on patients and their treatment. But there are many 
obstacles in the way of effective communication: Participants may use different terms to 
describe the same thing - a particular problem for patients who do not understand medical
terminology. Different participants frequently have different information needs and little time 
to filter information, so that no single report is truly adequate for all. And the different 
participants may rarely have time to meet, yet the care of a patient is shared and passed 
between them.

The information required by different participants is increasingly available in coded and 
structured forms - in the patient record, in drug databases, and in knowledge bases of medical  
terminology. For population studies (in epidemiology and administration), having data in a 
coded and structured form enables precise queries to be formulated and quantitative analyses 
to be done on large bodies of data. Yet when data are being used to convince, justify or 
explain, or to describe the status of a single individual, the more familiar medium of plain text 
and graphics may be more appropriate and effective: A focused coherent written report may 
be easier to deal with than the output from a set of database queries.

In healthcare, the evident need to translate between textual forms (human authored texts) and 
structured information has led to a large and continually growing body of research and 
development in natural language understanding \cite{fj92}.
In this article we consider the reverse problem - how textual documents may
be produced from structured data. In particular, we 
show how a range of current natural language generation techniques can be
used to produce from the same data, many different documents with different
content, terminology and style, and thereby help meet diverse information
needs within healthcare.

\section{A Brief Overview of Natural Language Generation}

Natural language generation (NLG) is concerned with automatically generating
texts in English (or other human languages) from computer-accessible data.
(See \cite{bh92} for a recent review.\footnote{A more detailed and
up to date survey is available online at web address
http://www.cse.ogi.edu/CSLU/HLTsurvey/HLTsurvey.html, chapter 4.})
NLG techniques range from the simplest report generation and mail merge
systems\footnote{Mail merge systems, available with most word processors,
allow information from a database to be incorporated into a text document in
simple ways, allowing, for example, mass production of personalised 
letters using information from a customer database. Simple IF-THEN
statements often allow different chunks of text to be output depending
on the current data.}, 
to sophisticated discourse and dialogue generation systems that reason about
the effect various forms and presentational structures will have on their
recipients. Simple mail merge techniques have been used in practice, as
well as in various experiments on whether personalisation of material used
in patient education \cite{osman94,campbell94,strecher94,meldrum94,brug96}
can increase its effectiveness. However, the problem 
with such techniques is that they are inflexible, allowing relatively little variation in the texts 
produced. Attempts at allowing wider variation may result in texts that are no longer coherent, 
requiring post-editing by a person. Improving such systems requires more knowledge of 
language; more sophisticated NLG systems and techniques exploit research on human 
language processing. For example, NLG systems can exploit linguistic theories about where 
pronouns can be used and what they can be used to refer to, to automatically choose between 
using a pronoun or a full noun phrase at a particular point in a text. This might partially avoid 
the need for human post-editing in a report generation system.

Natural language generation may be divided into stages. One proposed division
\cite{reiter94} is
\begin{itemize}
\item {\bf Text planning}:  The basic content of the text is selected for
the particular readership and organised coherently. Theories of text
organisation may be used to find a good ordering of information.
\item {\bf Sentence planning}: The information is split into sentences and
paragraphs, and appropriate use made of conjunctions, pronouns, etc.
\item {\bf  Realisation}: Grammatically correct sentences are produced.
A grammar of the language (e.g., English) may be used, and knowledge of
when different grammatical forms are appropriate. 
\end{itemize}

Consider the task of producing a summary of the structured information in a patient record. 
Text planning methods would be used to extract and select the relevant information and decide 
on the basic ordering of that information. For example, the most recent information might be 
selected, and ordered so that first all diagnoses and then all treatments are listed. Sentence 
planning methods would be used to divide that information into sentences, for example 
deciding to combine two separate facts (such as that the patient had a cough and that the 
patient has a sore throat) into one sentence. Finally, realisation methods
would be used to find a way to express the sentences in grammatical English,
for example as the sentence ``The patient has a cough and a sore throat.''

Not all systems will automate every stage. For example, current report generation and mail 
merge systems typically only provide for a simple form of text planning, while replacing 
sentence planning and realization with simple ``fill in the blank'' word strings. With respect to 
text planning, these systems allow information in a database or spreadsheet to control the 
selection of paragraph and sentence templates, but the basic form of each template must be 
provided by a human author: The system cannot automatically change the way facts are split 
into sentences or dynamically vary the grammatical forms or terminology used.

More complex methods of text planning, sentence planning and realization allow greater 
flexibility to be achieved: Terminology and style can be varied according to readership, and 
content selected to meet specific needs. The question that must be answered is how much 
flexibility is needed in a given application: if there is little variation in what the intended text is 
about or who is it for, then simple methods may suffice. However, the more variation in the 
intended audience, the more inherent complexity in the material to be presented, or the more 
the desire for communication to be a ``two-way'' interchange (i.e., a
dialogue) rather than a one-time delivery of information, the more
flexibility is needed. For many practical systems, only a limited amount
of flexibility may be required - for example, where the texts to be 
produced have a standard structure (e.g., SOAP notes \cite{weed69}), and the
process of content selection involves filling out this structure. In cases
where only a small number of basic sentence forms are regularly used, it
may be unnecessary to use a complex grammar to automatically generate
sentences, as basic sentence templates (e.g., ``Patient-X is suffering 
from Disease-Y'') may be provided for all the cases required. 

For a particular application we have to decide whether the additional flexibility we get from 
using the more sophisticated methods justifies the costs: the more
sophisticated methods do require more work in setting up for a particular
application \cite{rm93}. This will depend on the kind 
of flexibility required for that application.

\section{Example Applications}

Within healthcare natural language generation techniques have been applied in
a number of areas: generating explanations, advice and critiques in medical
expert systems
\cite{swartout81,lewin91,miller86,miller83,haim91,gertner94,gwc97,carberry97a};
generating reports, briefings, progress notes and discharge letters for health
professionals \cite{bullock94,bernauer91,camp93,abella95,dalal96a} and
generating explanatory materials for patients
\cite{migraine,buchanan92,decarolis96,binstead95,hirst97}. This review
considers each of these areas, considering the applicability of different
language generation techniques in each.

\subsection{Generating Expert System Explanations and Critiques}

If the recommendations of an expert system or decision support system are to be understood 
and assessed by health professionals, then some explanation of the reasoning or rationale 
behind the recommendation should be available. Generating such explanations is an NLG task 
- a text is generated from computer-accessible data concerning the systems reasoning.

In the earliest work on explanation in rule-based expert systems very simple NLG methods 
were viewed as sufficient, as the input was very constrained (just the trace of the system's 
reasoning) and relatively little variability in the output was required. Text could be generated 
to explain how the system reached a conclusion or why the system was asking a particular 
question. The text consisted of an English sentence whose clauses corresponded to the 
antecedent and consequent clauses of the rule from which a conclusion was drawn or that 
motivated a particular question being asked. No text planning was necessary, as only a single 
sentence was produced, and no sentence planning was necessary, as what was included in the 
sentence to be generated was fixed by the set of clauses in the rule.
Realization consisted of stringing together text using simple templates
associated with expert system rules or primitive functions. Early systems
such as MYCIN showed how far this basic approach could be taken \cite{scott84}.

The explanations produced using these simple methods are far from ideal.
However, improving them requires as a basis, a richer source of information  -
a simple rule trace will often miss key information that should be included
in order for an explanation to be convincing. Swartout \cite{swartout81}
attempted to address this issue by looking at how the underlying 
rationale behind rules may be preserved and used in explanation. Although only simple 
generation methods were used, this work is important in reminding us that one cannot get a 
good text without good data, and that to improve the output of a text generation system the 
first thing needed may be to improve the input.

Explanations were based on richer input in proto-type systems developed by
Langlotz \cite{langlotz87,langlotz88,langlotz:phd} and by Jimison
\cite{jimison90,jimison92}. Langlotz's QxQ system was developed to
demonstrate that the clinical use of quantitative decision models could be
facilitated through generating non-quantitative explanations of their
results - explaining which decision option was preferable, the basis for
that preference, and the sensitivity of that preference to uncertainty in
the relevant probabilities - using both text and graphics.

In producing its explanations, QxQ used a basic form of text planning: it had a small set of 
strategies modeled on the form and content of published medical decision analyses that were 
viewed as presenting persuasive arguments. QxQ used heuristic rules to select strategies that 
effectively used the available data to argue for the results of the model, and then merged the 
data with the selected frameworks to produce symbolic expressions that justify the difference 
in expected utility. QxQ's explanations required no sentence planning, as the type of material 
to be included in each sentence was specified in the strategies, and it used the same realization 
method as in the earlier MYCIN work. Since QxQ's explanations were longer and more 
complex than MYCIN's, more of the problems of inflexible realization were apparent in QxQ's 
explanations. For example, in explaining why intermittent pneumatic
compression prophylaxis is strongly indicated in a case of deep vein
thrombosis, QxQ produced the sentence:
\begin{quote}
``The decision is supported by the fact that the probability of deep vein
thrombosis with no prophylaxis is greater than the probability of deep vein
thrombosis with intermittent pneumatic compression prophylaxis.''
\end{quote}
If QxQ's realization strategies could make use of demonstrative pronouns and
contextual abbreviations, this sentence could be more simply realised as:
\begin{quote}
``This decision is supported by the fact that the probability of deep vein
thrombosis with no prophylaxis is greater than that with IPC prophylaxis.''
\end{quote}

Jimison's system produced explanations that incorporated patient-specific
characteristics (such as their occupation, leisure activities and past
experience of pain), as well as clinical factors \cite{jimison90,jimison92}.
These personal characteristics influence patient preferences for different
outcomes, and hence the choice of treatment. Jimison observed that texts
which present decision models implicitly refer to additional variables such
as these, yet these don't appear in the decision model. She showed how
adding such variables explicitly to the model, along with distributions 
on their range of possible values, could be used to produce explanations
that could contrast the recommendation for a specific patient with that for
the typical or generic patient. This allowed the physician who was given
the explanation to understand the importance of each variable to the
particular decision. The methods used for producing texts were similar to that
used in QxQ. While both QxQ and Jimison's system remained prototypes, both
stand as significant proofs of concept.

Because QxQ had only a small set of presentation strategies, it had no need for more complex 
text planning - for example, to mediate between strategies when more than one applied. More 
flexible and richer explanations may be generated by treating text or explanation planning as an 
independent problem. This is discussed by Moore \cite{moore94} (although
not for a medical domain). A simple example of independent text planning
in a medical expert system is HF-EXPLAIN \cite{lewin91}, which generates
explanations for a heart failure expert system based on a causal model. Rather 
than merely providing a trace of the causal processes, the system tries to
follow the typical structure of human explanations in this domain by making
use of simple schemata.

Turning from the generation of explanations to that of critiques, a
critiquing expert system comments on the user's suggestions rather than
generating its own. Such a system most therefore generate a coherent
critique based on its analysis of the user's proposal. For example,
ATTENDING analyses the risks and benefits associated with a proposed
anaesthesia plan, and generates an English critique \cite{miller86,miller83}.
The basic content and structure of this 
critique is fairly rigid - there is no explicit text planning stage. However, the way a critique is 
expressed requires flexibility because of the complexity of ATTENDING's analysis of the 
user's plan: simple template-based approaches would not be sufficiently flexible to produce 
fluent output. ATTENDING uses a slightly more complex realisation method for its 
generation system (PROSENET) based on an augmented transition network labelled with 
fragments of English. Traversing a particular network should result in a grammatically correct 
utterance whose details depend on contextual factors and details of the input. An example 
fragment of output is given in Figure 1.

\begin{figure}
\begin{center}
\framebox{\parbox{5.5in}{
Although intubation of this patient was not proposed, it is clearly
desirable. Not intubating this patient would have the risk of aspiration. 

\smallskip

Looking at the other aspects of the proposed plan, for a patient with
chronic renal failure, Curare is a reasonable selection since it is
reliably metabolized by the liver, and Halothane is a good choice since
it has no nephrotoxicity.}}
\end{center}
\caption{Example Text from ATTENDING critiquing system}
\label{attending:fig}
\end{figure}

Recent work on medical decision support has started to consider the issue of
how  evidence-based guidelines may be made widely available. In a
computer-based guideline system, although the underlying reasoning may be
simple, there is still the question of how best to present the guidelines
(and patient specific advice) to the health professional. This has been 
explored by Barnes and Barnett \cite{barnes95} and Day et al \cite{day95}.
Barnes and Barnett, for example, use an approach similar to PROSENET to
produce coherent patient specific guidelines.

\begin{figure}
\begin{center}
\begin{tabular}{p{6.0in}}
\underline{Original Critiques Produced by TraumaTIQ:} \\
*Caution: get a chest x-ray immediately to rule out a simple right
pneumothorax.
\\ 
 *Caution: get a chest x-ray immediately to rule out a simple right
hemothorax.
\\ 
*Do not perform local visual exploration of all abdominal wounds
until after getting a chest x-ray.  The outcome of the latter may
affect the need to do the former.
\\ 
*Please get a chest x-ray before performing local visual exploration
of all abdominal wounds because it has a higher priority.
\\ 
\end{tabular}
\end{center}
\begin{center}
\begin{tabular}{p{6.0in}}
\underline{Merged message:}
\\

*Caution: get a chest x-ray to rule out a simple right
pneumothorax and rule out a simple right hemothorax, and use the
results of the chest x-ray to decide whether or not to perform local
visual exploration of all abdominal wounds.
\end{tabular}
\end{center}
\caption{Critiques Produced by TraumaGEN}
\label{TToutput-1}
\rule{6.0in}{.010in}
\end{figure}

Where many guidelines are simultaneously active however, such simple
presentational methods may not be sufficient. In such cases, a text planner
must be able to take an arbitrary set of communicative goals, each relating
to a different piece of advice, and produce text that expresses the entire
set both concisely and coherently. This is done in TraumaGEN
\cite{carberry97a}, which has been designed to produce coherent critiques
in initial definitive management of multiple trauma. TraumaGEN works on the
output of TraumaTIQ \cite{gertner94,gwc97}, which produces individual
critiques of physician orders (or the lack thereof) based on what it infers
to be the physician's current plan and on the recommendations of its
associated expert system, TraumAID \cite{wrc92}. TraumaGEN addresses the
problem that, while in isolation each of TraumaTIQ's noted critiques may
effectively warn a physician about a problem, usually several problems
are detected simultaneously, producing multiple critiques whose aggregation
can be confusing (Figure 2). TraumaGEN takes the set of individual critiques
and integrates them into a more concise, coherent, and thereby more effective,
form.

\subsection{Generating Patient Information Materials}

There are many reasons why patients should be given better information:  to reduce patient 
anxiety; to enable patients to share in management decisions; to enable and encourage them to 
change their behaviour (e.g., to stop smoking); to enable and encourage them to comply with 
treatment; to enable them to manage chronic conditions; and simply to increase patient 
satisfaction. Improved patient education through better information has the potential to save 
large amounts of time and money, as well as lead to more satisfied and healthier patients. 
There is a large literature on the subject, but a good overview is given by
Ley \cite{ley88}.

Currently, patient education is largely provided through verbal interaction with health 
professionals, and through leaflets, posters and other printed material. However, health 
professionals have limited time (and are not always good communicators), and generic leaflets 
are impersonal and unspecific, not addressing a particular patient's particular needs. Patient 
education researchers recognise the need for more personalised materials
\cite{marshall84}, and simple mail merge techniques have been used effectively
to produce personalised leaflets \cite{osman94,campbell94,strecher94} which 
appear more effective than general ones. Simple interactive systems have
also been shown to be acceptable to patients \cite{jones93,gillespie93} -
patients may be less embarrassed asking questions of a 
computer than they are of their doctor \cite{barwani97}.

Although it is possible to produce personalised materials (leaflets and interactive systems) 
using fairly simple techniques, it is an area where NLG techniques can lead to more flexible 
systems and more coherent, fluent texts. Several recent projects (Migraine, Piglit, OPADE and 
HealthDoc) have explored this. In each, a text generation system has been used to produce 
leaflets or interactive materials, based on a combination of (a) data from drug databases and/or 
medical knowledge bases, and (b) data on the specific patient. Information is selected to be 
included in the text, based on the patient's needs, including information
specific to that patient.

The Migraine project \cite{migraine,buchanan92} is concerned with generating
interactive materials for migraine patients. Screens of text are generated
using a NLG system, with the user able to ask follow-up questions via a
mixture of hypertext and menu selection. A fairly sophisticated text planner 
\cite{moore94} is used to select content appropriate to each individual.
The approach allows responses to later questions to be answered in the
context of previous replies, referring back to these earlier responses and
avoiding excessive repetition \cite{carenini93a}. The medical knowledge
base required was constructed using the UMLS semantic network
\cite{carenini93b,lindberg93}, and patient data obtained through an 
online interview.

Text planning in Migraine proceeds by using text planning operators to expand goals into 
subgoals, depending on constraints, until a subgoal can be conveyed using a simple phrase or 
sentence. A sample text planning operator from Migraine is given in Figure 3.
It states that one way to alleviate a female pre-menopausal patient's fears
about the disease in question is to get them to believe that the disease
improves after menopause and with aging. Given such a goal 
of alleviating patient fears, this plan operator could be used to expand that
goal for an appropriate patient, eventually producing a phrase or sentence
aimed at getting them to believe 
that the disease improves after menopause and one aimed at getting them to believe that the 
disease improves with aging. This goal-directed method of selecting text content has proved 
more flexible than methods used in mail merge and similar software.

\begin{figure}
\begin{verbatim}
 (define-text-plan-operator
     :name alleviate-fears-female-pre-menopausal
     :effect (alleviate-fears ?patient (forever ?disease))
     :constraints
             ((female ?patient)
              (pre-menopausal ?patient)
              (not (in-patient-history? estrogens))
     :nucleus ((BEL ?patient 
                 (improve ?disease (after menopause)))
               (BEL ?patient
                 (improve ?disease aging)))))
\end{verbatim}
\caption{Example Text Planning Operator in Migraine}
\label{migraine:fig}
\end{figure}

Piglit \cite{binstead95} was similar to Migraine in many ways, using text
planning methods to generate 
hypertext explanations for diabetes patients. Piglit, however, had closer links with the patient 
record, enabling the patient to explore topics mentioned in their record, while the record was 
used in turn to determine how these topics should be explained, and to add personal 
reminders. Text planning was kept simple, with planning operators specifying the information 
that should be included when describing a particular class of medical concept (e.g., when 
describing a disease,  describe its symptoms and possible treatments). Information relevant to a 
specific patient could be added (e.g., if the patient has this disease, describe its onset and how 
it is being treated).  The medical knowledge base was constructed based on the organisation of 
concepts in the Read coding scheme (used by the National Health Service in the UK) but has 
since been modified to use the GRAIL representation language from the GALEN medical 
terminology project \cite{rector95}. Figure 4 gives an example text from the
original system (words in bold font can be clicked on for further information).

\begin{figure}
\rule{6.5in}{.010in}
{\large BEZAFIBRATE}

\medskip

Bezafibrate is a {\bf cardiovascular drug} which reduces the amount of some
kinds of fat in the bloodstream. According to {\bf your record} you are
currently undergoing this treatment. It is often used to treat
{\bf hyperlipidaemia}. It could have some side effects, in particular
{\bf nausea}. Your prescription of bezafibrate comes in 200 mg 
tablets. It is to be taken three times each day. 

\medskip
\begin{center}
YOUR RECORD~~~~~~~~~~MORE INFO~~~~~~~~~~HELP~~~~~~~~~~Discuss with doctor?
\end{center}
\rule{6.5in}{.010in}
\caption{Example Text from the PIGLIT system}
\label{piglit:fig}
\end{figure}

Both Piglit and Migraine use simple surface realisation based on selecting
from human-authored templates (e.g., ``The symptoms of Disease-X are
Symptoms-Y'') and filling in with specific data. They have both resulted
in working systems that have been evaluated with a small number of patients.
A much larger scale randomised trial is now in progress looking at 
the benefits of personalised information for cancer patients, using a system
based on Piglit \cite{jones96}. A recognised problem with both systems is the
effort required to author the knowledge 
base (containing the general medical information that may be communicated to the patient).  
However, once a suitable knowledge base is constructed then it appears easier to maintain it  
(as medical knowledge changes) than to maintain materials in a textual form. This is because  a 
particular fact (say, the drug recommended for a particular disease) will only appear once in 
the knowledge base, but may be mentioned many times in textual documents, using many 
different surface forms.

The OPADE project looked at generating personalised leaflets about drugs
\cite{decarolis96}. As in Piglit 
and Migraine, the emphasis was on text planning. A notable feature of OPADE was that the 
knowledge to be communicated all came from existing resources (a drug database and the 
prescription), so authoring or adapting a knowledge base was not required. OPADE also 
attempted to address the problem of the potential conflict between what the doctor wants to 
communicate and what the patient wants to know \cite{rosis95}.

The most recent project in this area is the HealthDoc project \cite{hirst97}.
HealthDoc takes an 
unusual approach to text generation. Rather than generating text from a knowledge base, 
HealthDoc starts with a master document. Generation involves selecting from this master 
document, and repairing the result to produce a coherent document.  For example, when 
material is cut from the master text, pronouns (e.g., ``it'') may be left
without a referent. Repairing a text might involve replacing one or more of
these pronouns with a full noun phrase (e.g., ``the angina pectoris'').

The HealthDoc approach thus places most of its effort at the sentence planning stage, and may 
prove an effective practical technique for applications where there is a relatively small amount 
of material to be selected from, and where existing report generation or mail merge software 
results in texts that require significant post-editing by a human.

The above systems also differ in the techniques that were used to elicit system requirements - 
the kind of material patients require, and how material can best be adapted for an individual 
patient. In the MIGRAINE project, ethnographic studies were used with
patients \cite{forsythe95}. In OPADE, surveys and questionnaires were used
to find which topics patients considered important, and what ordering of
information was prefered \cite{berry95}, as well as interviews with 
health professionals, who were asked to explain topics to hypothetical
patients \cite{decarolis96}. Piglit \cite{binstead95} used (primarily)
questionnaire-based feedback on early prototypes, but also interviews with
health professionals.

\subsection{Generating Reports and Progress Notes}

Another area in which NLG techniques have been tried experimentally in healthcare is in the 
generation of discharge summaries and progress notes. When the care of a patient is shared 
among or passed on to other health professionals, it is important that a clear and accurate 
account is presented of the current state of the illness and treatment. Yet producing such 
accounts is time consuming \cite{hohn95}, and the required clarity demands
both fluency and coherence \cite{quaak86}. Computer-based patient record
systems may already have simple report generation facilities built in,
which can be used to assist the healthcare professional. Because the resulting 
output may lack fluency, requiring significant post-editing, several
researchers in medical informatics have looked at ways of generating better
reports \cite{bullock94,bernauer91,camp93,trace93}.

Generating better reports does not appear to require sophisticated text planning methods, as 
the structure of a progress note or discharge summary is generally fairly constrained. Indeed, 
the requirement for a consistent, easy-to-scan format may mean that too much variability in 
the output is undesirable. However, merely stringing together data in a sequence of short 
sentences or clauses will result in a report that lacks fluency and is overly verbose. Here, what 
is required are sentence planning techniques, especially ones that deal with aggregation (i.e. 
combining and merging clauses into concise sentences) and with pronominalisation, both of 
which reduce repetition - hence reducing verbosity - and increase coherence, by bringing 
together related material. Simple methods can normally be used for realisation, as frequently 
only a small number of basic sentence forms are needed.

The IVORY system \cite{camp93}, for example, was designed to generate
textual progress notes given 
data entered by the physician. User-centered design methods were used to try and ensure that 
the system met the needs of physicians, and the format of the notes followed that used by 
physicians when writing progress notes by hand. In this case the SOAP format was used, 
where first subjective data is given (patient's reported symptoms), then objective data 
(physician's observations and measurements), then the physicians assessment, and finally the 
physician's plan for treatment or further tests. While text planning in IVORY simply involved 
filling out the basic SOAP template with the relevant data, significant attention was paid to 
sentence planning, particularly combining clauses as a way of avoiding repetition. As a simple 
example, rather than generating ``Three day history of cough. Three day
history of sore throat.'' IVORY would generate ``Three day history of cough
and sore throat.''

An example fragment from a progress note generated by IVORY is given in
Figure 5. IVORY 
does not generate fully grammatical English sentences, but rather abbreviated phrases, similar 
to those used by many physicians and nurses. (The style is, in fact,  rather standard 
"telegraphic" American English, which can be observed in a wide-range of applications and is 
not specific to health care.) The structured progress note form is preferred by many health 
professionals, allowing easy access to relevant parts of the report. As the telegraphic style 
used in IVORY employs rather short constructions, a sophisticated grammar-based realisation 
component was not seen to be required.

\begin{figure}
\rule{6.5in}{.010in}

\noindent
{\large {\bf SUBJECTIVE:}}

\medskip

\noindent
{\bf Constitutional}:\\
Mild frontal headaches for the last 2 days. No fever and no chills.

\smallskip

\noindent
{\bf ENT}:\\
Constant, moderate sore throat for the last 1 day. The sore throat is
worsening. No nasal discharge.

\medskip

\noindent
{\large {\bf OBJECTIVE}}:

\smallskip

\noindent
{\bf Vital Signs}:\\
Oral temp: 99.8 F. Right brachial pulse: 89 sitting. Right upper arm bp 130/85 
sitting....

\rule{6.5in}{.010in}
\caption{Fraction of Progress Note Generated by IVORY}
\label{IVORY:fig}
\end{figure}

A similar system has been developed recently \cite{bullock94} for generating
summaries from the Pen\&Pad patient record system \cite{rector91}. While most
work in this area focuses on the specific needs of the application, this
work specifically considers NLG issues. Text planning is based 
on simple schemas \cite{mckeown85} which provide a flexible way to structure
the text. Sentence planning includes aggregation and pronominalisation
methods, and a realisation module is based on a simple grammar, producing
grammatically well-formed output (illustrated in Figure 6).

\begin{figure}
\begin{center}
\framebox{\parbox{5.5in}{The angine pectoris is progressive and moderate. It has an aching,
burning, possibly intermittent character, is aggravated by cold and movement
and relieved by rest. There is a cough and no breathlessness.

\smallskip

The cough is improving and mild. It has a dry, non-productive character and a 
bovine, harsh sound character. It is aggravated by dust and smoking and
relieved by rest}}
\end{center}
\caption{Example Report from Pen \& Pad Reporter}
\label{pp:fig}
\end{figure}

Other report generation applications have used more sophisticated realisation
modules, corresponding to more variability in the input. For example,
Abella et al \cite{abella95}, describe a system for generating reports on
radiographs, using a fairly sophisticated ``off-the shelf''
sentence planning and realisation system (FUF \cite{elhadad92}) to produce
sentences. Bernauer et al \cite{bernauer91} argue that for producing reports
on bone scan studies, the variety of possible kinds of 
observations requires a relatively sophisticated realisation module;
they use one developed specially for their task. It appears that for generating medical reports there is no one 
generation architecture perfect for every application, but each specific application may make 
different demands and require different sorts of flexibility and variability to be supported.

The work mentioned above has been concerned with generating a written textual report. 
However, some information can be conveyed more effectively through graphics, while the use 
of spoken (rather than written) text can allow someone to attend to both text and graphics (or 
to both text and the world outside) simultaneously. This has led to a recent interest 
inmultimedia generation. To generate a multimedia presentation from data involves selecting 
the appropriate modality (text, graphics etc) for communication, and coordinating the 
presentation so, for example, images are referred to in the text. Recent work at Columbia \cite{dalal96a} has considered how multimedia post-operative
briefings can be generated that meet the different needs of nurse and
physician.

\subsection{Generating Descriptions of Medical Concepts}

In most of the above applications there is a need for generating descriptions of medical 
concepts given a standard code (e.g., SNOMED, ICD-9) for that concept. The complexity of 
this problem depends on the coding scheme used. In a system (such as ICD-9) where a 
separate code is given for every distinguishable concept, then a simple mapping of code to 
preferred English phrase may be possible. But in more complex compositional schemes, such 
as that used in the GALEN terminology server \cite{rector95}, generating the
appropriate English phrase for a compositional concept becomes more
difficult \cite{solomon96,wagner95}. However, although the 
generation of noun phrases is more complex, the approach can pay off if
multilingual output is required \cite{elhadad92}. While the 1997 version of
the UMLS \cite{lindberg93} correlates terminologies in Spanish, German,
Potuguese and French with those in English, the mapping of terms is only
partial - there are only translations available for all these languages for
MeSH terms. For these languages and others, entering a preferred phrase
by hand for each uncovered concept will require extensive effort. In a
compositional scheme, on the other hand, much can be achieved 
based on combining words from a relatively simple lexicon.

\section{Discussion}

The above systems have used a wide range of natural language generation (NLG)
techniques, from the very simple to the complex. Some have emphasised text
planning
\cite{swartout81,lewin91,carberry97a,migraine,buchanan92,decarolis96,binstead95},
others sentence planning \cite{bullock94,camp93,hirst97} or realisation
\cite{miller86,miller83,wagner95}. In most cases those using 
more sophisticated techniques
\cite{swartout81,lewin91,miller86,miller83,haim91,carberry97a,bullock94,dalal96a,migraine,buchanan92,decarolis96,binstead95,hirst97,wagner95}
are research prototypes, which, while aiming to address a real need, have
not been used in practice beyond small scale 
evaluations. Here we briefly consider both where more advanced NLG techniques are likely to 
be worthwhile in practice, and which methods are likely to be most useful.

Sometimes the costs of using advanced NLG techniques may outweigh the
benefits \cite{weed69}: For 
example, if using them requires authoring a special purpose knowledge base whose life-span is 
short, then unless the techniques allow one to serve a significantly larger population that one 
otherwise could, then simple techniques, such as those used in mail merge systems, may be 
adequate. One useful compromise that can be seen even in the research systems discussed 
earlier, may be to use advanced techniques for part of the process and simpler techniques for 
the rest. For example, simple text planning can be combined with sophisticated sentence 
planning and realisation methods if good fluency is required, or depending on context. 
Alternatively, simple fill-in-the-blanks sentence templates can be combined with sophisticated 
text planning methods if the selecting and structuring of the content is quite complex.

Clearly, whatever techniques are used, for a system to be used in practice it is important that it 
is integrated both with existing clinical systems (patient records, drug databases etc),  and also 
with existing practice. This point has been frequently made for medical expert systems and 
applies equally to NLG systems used in healthcare. A system that gave
``added value'' to an 
existing patient record system would be more persuasive than a stand-alone system requiring 
separate or idiosyncratic data entry. Research prototypes may not go as far as  using actual 
patient record systems, but should at least take account of the medical coding schemes in use 
in such systems. 

\section{Conclusion}

Healthcare communication is an area where there is a real need for the generation of textual 
reports and explanatory materials from structured data. Natural language generation is a  
rapidly maturing field. As techniques become better understood and more
``off-the-shelf'' tools become readily available, NLG offers real potential
for better healthcare communication, 
increasing the flexibility and adaptability of systems and the fluency of output texts. There will 
always be a cost associated with using more sophisticated techniques, and these costs must be 
weighed against the benefits. However, intermediate techniques, using simpler techniques for 
part of the process, may provide ways to get maximum benefit for a particular application with 
minimal cost. The appropriate techniques will depend on the kind of flexibility and the style of 
text required for the particular problem. 

\section{Acknowledgements}

This work has been supported in part by the National Library of Medicine,
under grants R01 LM05764-03 and RO1 LM06325-01, and the Engineering and
Physical Sciences Research Council, under grant GR/K55271.

\end{document}